\documentclass{article}

\usepackage{PRIMEarxiv}

\usepackage[utf8]{inputenc} % allow utf-8 input
\usepackage[T1]{fontenc}    % use 8-bit T1 fonts
\usepackage{hyperref}       % hyperlinks
\usepackage{url}            % simple URL typesetting
\usepackage{booktabs}       % professional-quality tables
\usepackage{amsfonts}       % blackboard math symbols
\usepackage{nicefrac}       % compact symbols for 1/2, etc.
\usepackage{microtype}      % microtypography
\usepackage{lipsum}
\usepackage{fancyhdr}       % header
\usepackage{nicematrix}
\usepackage{graphicx}       % graphics
\graphicspath{{media/}}     % organize your images and other figures under media/ folder

\title{Evaluation Metrics for Text Data Augmentation in NLP
}

\author{
  Marcellus Amadeus, William Alberto Cruz Castañeda \\
  Alana AI Research \\
  Brazil \\
  \texttt{\{marcellus, william.cruz\}@alana.ai} \\
}

\begin{document}
\maketitle

\begin{abstract}
Recent surveys on data augmentation for natural language processing have reported different techniques and advancements in the field. Several frameworks, tools, and repositories promote the implementation of text data augmentation pipelines. However, a lack of evaluation criteria and standards for method comparison due to different tasks, metrics, datasets, architectures, and experimental settings makes comparisons meaningless.  Also, a lack of methods unification exists and text data augmentation research would benefit from unified metrics to compare different augmentation methods. Thus, academics and the industry endeavor relevant evaluation metrics for text data augmentation techniques. The contribution of this work is to provide a taxonomy of evaluation metrics for text augmentation methods and serve as a direction for a unified benchmark. The proposed taxonomy organizes categories that include tools for implementation and metrics calculation. Finally, with this study, we intend to present opportunities to explore the unification and standardization of text data augmentation metrics
\end{abstract}

% keywords can be removed
\keywords{text data augmentation \and natural language processing \and unified metrics \and taxonomy}

\section{Introduction}
Recent surveys on data augmentation (DA) for natural language processing (NLP) have reported different classifications, techniques, and advancements in the field. An overview exposed by \cite{bayer2022} presents more than 100 DA methods for text classification organized into 12 different groups and provides state-of-the-art references that expound what kind of methods are promising by relating them to each other. In addition, a list of frameworks and practical tools promotes DA implementation pipelines and raises the necessity of comprehensive evaluation criteria and standards for method comparison. The survey of \cite{steven2021} describes methodologies and DA techniques for NLP categorized into rule-based, example interpolation-based, or model-based. It discussed the usefulness of DA techniques, including low-resource languages, mitigating bias, fixing class imbalance, few-shot learning, and adversarial examples. Moreover, a lack of methods unification exists, and popular ones are present in an auxiliary fashion. Further, text DA research would benefit from unified benchmark tasks and datasets to compare different augmentation techniques.

To stimulate the use of strategies for selecting and constructing text DA techniques, \cite{bohan2022} divide into three categories according to the diversity of their augmented data: paraphrasing, noising, and sampling. Paraphrasing-based methods are thesauruses, semantic embeddings, language models, rules, machine translation, and model generation. The noising-based methods are swapping, deletion, insertion, and substitution. The sampling-based methods are rules, non-pretrained models, pretrained models, self-training, and mixups. For industrial applications, \cite{liu2020} presents text DA techniques such as back-translation, lexical substitution, and conditional DA. \cite{shorten2021} presents a list of augmentation frameworks for text data implementing symbolic or neural methods. Symbolic methods use rules or discrete data structures to form synthetic examples. Alternatively, the neural method implements a deep neural network trained on a different task to augment data. However, to compare the performance of DA techniques with its characteristics, the experimental settings make direct comparisons meaningless. Thus, the work contribution is to provide a taxonomy of evaluation metrics for text augmentation methods and serves as a direction for a unified benchmark. This work exposes the metrics found in the NLP literature between 2018 to 2023. The taxonomy includes tools and repositories to implement and calculate the metrics. Following is described each of the categories of the taxonomy. Finally, with this study, we intend to expose opportunities to explore the unification and standardization of DA metrics.

\section{Recent Development on Data Augmentation for NLP}
\label{sec:append-how-prod}

The problem of generating labeled sentences with generative adversarial networks (GANs) for DA is studied by \cite{yang2018}. Combine reinforcement learning (RL), GANs, and recurrent neural networks (RNN) to build a category sentence GAN (CS-GAN) model. The proposed model generates categorized sentences that increase the original dataset and improve its generalization capability during supervised training. Setup and training with conditional GAN (cGAN) are described by \cite{gupta2019} for DA in sentiment classification. Moreover, it addressed the training methodology and the architecture of the cGAN on low-resource sentiment analysis datasets. A conditional BERT with contextual augmentation is proposed by \cite{wu2018} for labeled sentences. BERT allows sentence augmentation without breaking the label compatibility and introducing a new conditional masked language model task. The proposed method is an unidirectional language model and can be applied to enhance contextual augmentation. Working with hate speech classification, \cite{rizos2019} proposes three techniques to reduce the degree of class imbalance and maximize the amount of information extracted from limited resources. The first technique is synonym replacement on word embedding vector closeness. The second uses warping techniques of the word tokens along the sequence or class-conditional, and the third uses recurrent neural language generation. All the techniques follow the process of text acquisition, pre-processing, word embedding, 1D convolutional neural network (CNN), max pooling, gated recurrent unit/long short-term memory (LSTM), max pooling, dense layer, and softmax activation.

Inspired by the capabilities of pre-trained word embedding models, \cite{hailu2019} propose a word embedding-based parallel text augmentation technique, which increases the size of the limited text for low-resource neural machine translation (NMT). Improving NMT robustness, \cite{li2019} proposes DA techniques to extend limited noisy data, injecting synthetic noise samples and adversarial attacks. Results show that techniques could benefit from data injected with noise through manual transcriptions of spoken language. The automatic speech recognition (ASR) generated data don't improve robustness, while the human transcripts from speech proved helpful in translating noisy texts. Easy DA (EDA) is a set of techniques developed by \cite{wei2019} that boost performance on text classification tasks. EDA consists of four operations: synonym replacement, random insertion, random swap, and random deletion. Results show that EDA improves performance with CNN and RNN and demonstrates strong results for smaller datasets. It highlighted the lack of standardized DA in NLP by introducing a set of simple operations that serve as a baseline for future investigation. At the word and sentence level, \cite{yu2019} proposed the Hierarchical DA (HDA) method for augmenting texts. HDA implements an attention mechanism to distill important contents from hierarchical texts as summaries. As a complement, training in the hierarchical attention networks model allows attention values of all documents. The experiments reveal that HDA is a promising technique when compared with EDA. The proposed approach of \cite{huong2020} is based on text DA for product reviews of online opinions in the Vietnamese language to substitute large amounts of human-labeled data that is costly to obtain. The word embedding implements cosine distance for measuring the similarity. Stopwords are created manually based on their TF-IDF score. Techniques are applied to generate comments by random insertions, substitutions, word swapping, word deletion based on word2vec, and word replacement by synonyms or similar words for Vietnamese text. The method of \cite{shim2020} combines EDA and pseudo-labeling to increase the size of the initial train set. A pre-trained BERT model is used as a text classifier and fine-tuned the data.

\cite{verissimo2020} evaluates the influence of the parameters of a DA technique in a textual dataset and the use of neural networks for emotion analysis of Brazilian Portuguese texts. Implements a multi-lingual fine-tuning method, which applies the concept of transfer learning. The text augmentation approach consists of synonymous words instead of original ones without losing the sentence’s sense. The study of \cite{chen2020} presents MixText, a semi-supervised learning method with a DA technique called TMix. TMix takes two text instances and interpolates them in their corresponding hidden space. Since the combination is continuous, TMix has the potential to create an infinite amount of augmented samples and avoid overfitting.

Comparing different DA techniques is a non-trivial task that requires incorporating different architectures and model implementation. To address these challenges, \cite{qiu2020} develop EasyAug, a DA platform that provides several augmentation approaches. EasyAug incorporates all prominent textual DA approaches. Provide three methods: random resampling, word-level transformations, and VAE-based text generation models. \cite{feng2020} propose and evaluate various augmentation methods, including GPT-2 fine-tunned version. Explore semantic text exchange, synthetic noise, synonym replacement, hyponym replacement, hypernym replacement, and random insertion. The quality of the generated text is similar to the original data. \cite{kumar2020} study GPT-2, BERT, and BART for conditional DA. These methods integrate text content manipulation as co-training the data generator and augmentation techniques combined with latent space augmentation.

Data Boost, described by \cite{liu-2020}, is an RL-guided text DA framework built on off-the-shelf GPT-2. Requires collecting extra data and training a task-specific language model from scratch. The generated samples serve as augmentation data similar to the original. A nonlinear Mixup, proposed by \cite{guo_2020}, addresses limitations in DA with Mixup. The nonlinear mixing policy mixes sample pairs on the word embedding level and deploys a separate mixing policy in the word dimensions of a given sentence. \cite{xie2020} presents, Unsupervised DA (UDA), a method for effectively noise unlabeled examples. This procedure forces the model to be insensitive to the noise and hence smoother concerning changes in the hidden space. From another perspective, minimizing the consistency loss propagates label information from labeled examples to unlabeled ones. For text, UDA combines well with representation learning as BERT. The contribution of \cite{marivate2020} is a short survey of several DA methods. The work categorizes text augmentation techniques in text source and text representation. Besides, an alternative way of augmenting low-resource language is with unsupervised word embedding models. The mixup-transformer architecture proposed by \cite{sun2020} applies mixup into transformer-based pre-trained models. Unlike static mixup, the architecture can dynamically construct inputs for text classification. Multi-granularity text-oriented DA technologies proposed by \cite{sunxiao2020} generate large-scale artificial data into a hybrid neural network model. The model combined two channels at the feature level and was co-trained to obtain the final feature representation for the input sentences.
 
The DA technique introduced by \cite{yuan2020} avoids involving noisy data with a keywords-oriented DA (KDA) method that extracts keywords based on category labels and augments them based on the keywords. The proposed technique selects the essential information and expands the selected data. KDA is inspired by the EDA approach, extracting the keywords from the text and then applying the DA technique to the data. A text augmentation method, presented by \cite{shyang2020}, scales the size of an emotion corpus by including only relevant positive examples. Tweets labeled with happiness are used as gold standard seeds to augment the training data to include similar distant supervision. \cite{thakur2021} presents a technique called Augmented SBERT, where the BERT cross-encoder label input pairs to fine-tune the bi-encoder and augment the training data. Based on LSTM networks, \cite{giannis2021} proposes three text DA techniques: permutation, antonym, and negation. Sentence permutations expand an initial dataset while retaining the statistical properties, applied multiple times per dataset without duplicating entries, thus avoiding overfitting. Antonym replacement and negation insertion reverse the classification of each augmented example but require mutually exclusive classes.

\cite{carrasco2021} generate sentences for five regional Arabic dialects. The objective is to overcome the scarcity of richly annotated dialectal Arabic datasets. The dataset has five regions, represented by classes corresponding to a different dialect. The Sentimental GAN (SentiGAN) model is used to populate low-resource dialects. However, one generator per class instead of multiple generators has five generators/discriminators working independently on a specific dialect without cross-group mixing. The main obstacle in training neural models for data-to-text generation is the lack of training data. To address this problem, \cite{chang2021} proposes a few-shot approach to augment the data in three ways. Train to generate new text samples based on replacing specific values with alternative ones from the same category. Based on GPT-2, generate text samples and propose an automatic method for pairing the new text samples with data samples.

Most text GANs are unstable in training due to inefficient generator optimization and rely on maximum likelihood pre-training. The work of \cite{lee2021} addresses the above problems by proposing an augmentation method with a Sentence Generator (SG) and Sentence Discriminator (SD) for Iterative Translation-based DA (ITDA). The ITDA SG provides universal multiple-language support by generating comprehensive augmented sentences through serial and parallel iterations of an existing translator. Implementing BERT for Persian tasks, \cite{varasteh2021} uses the pre-trained language model (PLM) ParsBERT to augment the input data and then use it for text classification. The augmentation enables the generalization of unseen and new words well. For patient outcomes prediction, \cite{lu_qiuhao_2021} proposes an architecture using fine-tuned GPT-2 for Medical text Augmentation (MedAug) to synthesize labeled text with the original training data. The study of \cite{shang_2021} proposes a sentiment analysis framework powered by two performance boosters: 1) a DA technique based on a transformer-based GAN and RoBERTa, and 2) an optimized BERT to improve the prediction accuracy. The GAN-based DA technique can generate high-quality synthetic samples to increase the size and diversity of the training set. \cite{liu__xiaomeng_2021} present an enhanced UDA (EUDA) by mixing DA strategies and using a problem-related prefilter. The DA strategies for UDA enhancement are back translation, TF-IDF, and EDA techniques. \cite{bae2021} compare a variational autoencoder (VAE) with EDA to increase the number of training samples for malware detection. \cite{rashid2021} modify the EDA technique with sophisticated text editing operations powered by masked language models to analyze the benefits or setbacks of creating a linguistical quality augmentation.

Virtual DA (VDA) is a general framework for fine-tuning PLM developed by \cite{zhou2021}. The idea is to generate DA at the embedding layer of PLM. To guarantee semantic relevance, consider a multinomial mixture of the original token embeddings as the augmented embedding. There are two advantages to this approach. First, the original token embeddings are the representation basis, and the augmented embeddings stay close to the existing embeddings. Second, the injected Gaussian noise generates variations for augmentations.

\cite{wilton2022} works with the DA technique to improve the performance of the BERT classification model for fake news in the Portuguese language. The DA technique consists of manipulating the dataset with modified copies or creating synthetically from existing data. The built process applies DA before translation takes place. \cite{yeong2022} proposes a text augmentation based on attention score (TABAS). The proposed approach recognizes that a criterion for selecting a replacement word rather than a random selection is necessary. The attention score processes only words with the same entity and part-of-speech tags to consider informational aspects. TABAS has two steps. The first step is dataset preparation, training an attention score, and building a word dictionary. In the second step are tokenized sentences in the dataset using the attention model and the word dictionary.

Studying the advantages and drawbacks of text augmentation methods \cite{abonizio_2022} evaluated strategies of text DA techniques on datasets for sentiment analysis and related tasks using four different classifiers. The proposed dual text DA strategy of \cite{bonthu2022} uses a mix of four pre/post-augmentation operations in two different settings on a classification model based on RNN. Considering a dynamic policy scheduling, \cite{shuokai2022} design a search space over augmentation policies by integrating generic augmentation operations. It adopted a population-based training method to search for the best augmentation schedule. Tailored Text Argumentation (TTA) proposed by \cite{feng2022} has two operations. The first is the probabilistic word sampling for synonym replacement based on the discriminative power and relevance of the word to sentiment. The second is the word identification and the zero masking or contextual replacement of these words. A two-step DA process developed by \cite{wadhwa2022} consists of a former stage with a method for preparing a comprehensive list of identity pairs with word embeddings. The next step prepares an identity pairs list to enhance the training instances by applying three simple operations (namely identity pair replacement, identity term blindness, and identity pair swap). Thus, an Identity Information DA method (IIDA) bolsters text classification models against unintended bias by automatically generating diverse instances. \cite{KimWoo2022} presents a DA using lexicalized probabilistic context-free grammar that generates augmented samples with diverse syntactic structures with plausible grammar. Another contribution is to the train-validation splitting methodologies, where the traditional splitting of training and validation sets is sub-optimal compared to our novel augmentation-based splitting strategies. \cite{saeedi2022} presents a combination of DA techniques to boost the performance of state-of-the-art transformer-based language models for Patronizing and Condescending Language (PCL) detection and multi-label PCL classification tasks. The approaches rely on fine-tuning pre-trained RoBERTa and GPT3 models with extra-enriched PCL datasets.

Inspired by AutoAugment, \cite{chu2022} proposes an automated DA for language processing and understanding with twofold novelties. First, an adaptive augmentation policy optimizes individual data points, sentences, or documents. Second, an efficient evaluation strategy designs training for auto augmentation policy networks. \cite{prakrankamanant2022} proposes an alternative DA technique to improve the robustness of poor tokenization by using multiple tokenizations. The developed technique for the Thai language uses different and possibly errorful tokenizations to enhance the Thai text classification.

NEO Natural Language DA (NEO-NDA) is a comprehensive tool to address data generation and rebalancing datasets proposed by \cite{ladeira2022} that work with multiple languages, create new samples, boost the performance of ML models, and handle multilingual DA. Offer a hybrid (rule-based and model-based) group of techniques for DA in seven different languages: English, French, German, Italian, Portuguese, Romanian, and Spanish. The study of \cite{wu2022} proposes a hybrid framework that implements a GAN and Shapley algorithm based on EDA to obtain balanced data for classifier training, generates an initial training sentence, and uses it as the seed data for the next GAN-based model. TreeMix is a compositional DA approach for natural language understanding proposed by \cite{zhang2022}. Leverages constituency parsing trees to decompose sentences into constituent sub-structures and the Mixup DA technique to recombine them to generate new sentences. Exploring the potential of using text-to-text language models, \cite{chenyan2022} proposes T5 and BART integration to construct a two-phase framework for augmentation. First, a fine-tuning phase where PLMs are well adapted to downstream classification. Second, a generation phase where the fine-tuned models create samples for performance lifting.

\cite{byeong2022} proposes DA with Generation and Modification (DAGAM), which combines DA techniques using a generation model (DAG) and character order changing (DAM) for a boosted performance. DAG uses a language generation model for DA. DAM exploits a psychological phenomenon called the word superiority effect on text data. Introduce three DA schemes that help to reduce the underfitting problems of large-scale language models. To support effective text classification, \cite{zhao2022} presents a plug-in DA framework (EPiDA). EPiDA employs two mechanisms to control the diversity and quality of augmented data: relative entropy maximization (REM) and conditional entropy minimization (CEM) for data generation control. REM enhances augmented data diversity, and CEM ensures semantic consistency.

\cite{kong2022} combines the advantages of Dropout and Mixup to develop a new approach for DA and regularization called DropMix to mitigate the overfitting problem in text learning. DOUBLEMIX, proposed by \cite{chen2022}, is a simple interpolation-based DA approach to improve the robustness of neural models in text classification by mixing up the original text with its perturbed variants in hidden space. \cite{vudang2022} implements several DA techniques in Korean corpora with PLMs (SKT-koBERT, SKT-GPT2, KrBERT, ENLIPLE-v2, KoELECTRA-Base-v3, BERT-Base).

\cite{yu2023} proposes a mixing-based text augmentation approach based on explainable artificial intelligence (XAI) to consider the importance of mixed-up words in labeling the augmented data. To improve the robustness of PLMs (BERT, DistilBERT), \cite{tang2023} proposes two DA techniques for text classification: cognate-based and antipode-based. A DA technique ChatAug proposed by \cite{dai2023}, based on ChatGPT, generates auxiliary samples for few-shot text classification. ChatGPT is trained on data with unparalleled linguistic richness and employs a reinforcement training process with large-scale human feedback, which endows the model with affinity to the naturalness of human language. The ChatAug rephrases each sentence in training into semantically different samples. \cite{kwon2023} proposes an XAI-based mix-up approach that explicitly derives the importance of manipulated words and reflects this importance in labeling augmented data.

\section{Taxonomy of Evaluation Metrics}

A taxonomy organizes categories that represent mappings of the text DA techniques. The result is shown in Figure 1 with ten categories as follows: human evaluation, machine translation quality, text generation quality (split up into novelty, diversity, and fluency), character n-gram matches; prediction quality for classification; datasets relationship; ASR Performance; training; language model robustness, and manipulated words importance. As a fundamental concern in this work, each metric in a category describes procedures, computational tools, or repositories to promote their calculation and implementation.

\begin{figure}[h]
\centering
\includegraphics[scale=0.8]{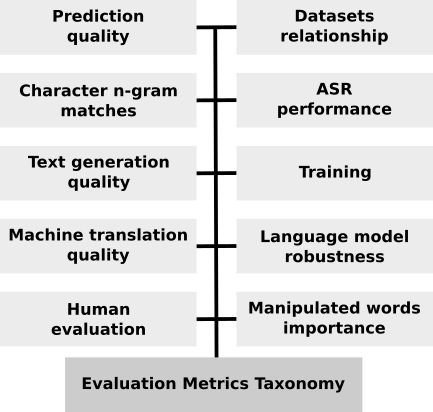}
\caption{Taxonomy of evaluation metrics for text augmentation methods.}\label{fig:1}
\end{figure}

\subsection{Human Evaluation}
This category checks the text generation quality from the human point of view. No computational implementation allows its use. The main objective is to obtain a fluency score and count of information missed and wrong information. Thus, the generated text maintains the information's completeness and correctness. Table \ref{tab:table_1} explains the procedure.

\begin{table}[h!]
\centering
\begin{tabular}{ll}
\hline
\textbf{Reference} & \textbf{Procedure}\\
\hline
\multicolumn{1}{m{1.5cm}}{\cite{chang2021}} &  \multicolumn{1}{m{10cm}}{The data and generated text are evaluated by individuals. For each data-text pair, the annotator is instructed to evaluate: a) if the text is fluent (score 0-5 with 5 being fully fluent), b) if it misses information contained in the source data, and c) if it includes wrong information.}\\
\hline
\end{tabular}
\caption{Human evaluation for text generation quality.}
\label{tab:table_1}
\end{table}

\subsection{Machine Translation Quality}
Human assessment is the best option for evaluating translation quality and is considered ground truth. However, as human evaluation is expensive, it is impossible to have every new output of the model evaluated by a group of individuals. Thus, this category assesses machine translation through the Bilingual Evaluation Understudy (BLEU) metric score. BLUE is a corpus-level metric based on the modified-gram precision measure with a length penalization for the candidate sentences. Table \ref{tab:table_2} presents the common computational implementations of this metric in Python language. Metrics 1, 2, 3, and 4 represent the precision of the n-gram sequence while calculating the BLEU score.

\begin{table}[h!]
\centering
\begin{tabular}{cc}
\hline
\textbf{Metric} & \textbf{Python-based Tools}\\
\hline
\multicolumn{1}{m{3cm}}{BLEU score, \break Macro BLEU-2} &  \multicolumn{1}{m{4.5cm}}{\quad \href{https://pypi.org/project/bleu/}{\includegraphics[scale=0.06]{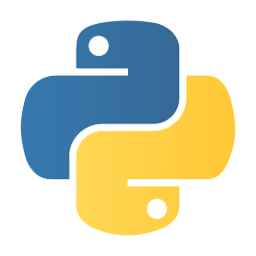}} \quad \href{https://huggingface.co/spaces/evaluate-metric/bleu}{\includegraphics[scale=0.1]{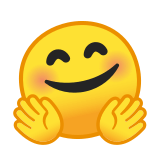}} \quad \href{https://www.nltk.org/api/nltk.translate.bleu_score.html}{NLTK}} \\
\hline
\end{tabular}
\caption{Metrics for machine translation quality involved in text data augmentation.}
\label{tab:table_2}
\end{table}

\subsection{Text Generation Quality}
This category joins five aspects required to define text generation quality metrics involved in text DA: novelty, diversity, fluency, semantic content preservation, and sentiment consistency. Novelty, an independent metric, verifies text consistency created by GANs and is not measured by the fluency or logic in the sentence. Ensure the generator creates new sentences instead of copying the ones in the dataset. No computational implementation exists and is defined as follows:
\[Novelty(S_i)=1-max\{\varphi(S_i,C_j)\}_{j=1}^{j=|C|} \]
where $C$ is the sentences set, $\varphi$ is the Jaccard similarity function and $S_i$ is a generated sentence. The diversity assesses the difference between generated and original sentences in the same dataset. The metrics to measure it are self-bleu, unique trigrams (UTR), type-token ratio (TTR), and rare words (RWORDS).  Self-BLEU evaluates the variety of sentences, measuring the BLEU score for each generated sentence by considering other generated sentences as references. UTR measures the ratio of unique to total trigrams in a population of generations, high UTR represents excellent diversity. TTR is the ratio of unique to total tokens in a piece of text and serves as a measure of intra-continuation diversity. The higher the TTR, the more varied the vocabulary in a continuation. In RWORDS, lower values indicate the usage of more rare words (less frequent in the corpus) and higher diversity.

Perplexity (PPL) is a metric for evaluating language models. Measure the probability of a sentence produced by the model trained on a dataset. With a lower perplexity value, the model predicts a sample better. Syntactic log-odds ratio (SLOR) modifies PPL by normalizing individual tokens. Higher SLOR represents higher fluency. SPELLCHECK measures two spelling-related metrics for synthetic noise. SPELLWORDS is the average number of misspelled words per continuation, and SPELLCHARS is the average number of character-level mistakes per continuation. Table \ref{tab:table_3} presents computational implementations in Python language for these metrics.

\begin{table}[h!]
\centering
\begin{tabular}{cc}
\hline
\textbf{Metric} & \textbf{Python-based Tools}\\
\hline
\multicolumn{1}{m{5cm}}{\textbf{Diversity} \break		 Self-BLEU, UTR, TTR, RWORDS} & \multicolumn{1}{m{6cm}}{\quad \href{https://pypi.org/project/fast-bleu/}{\includegraphics[scale=0.06]{images/python.png}} \quad \href{https://github.com/geek-ai/Texygen}{\includegraphics[scale=0.08]{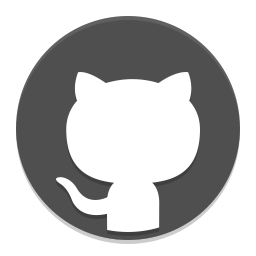}} \quad \href{https://pypi.org/project/lexicalrichness/}{\includegraphics[scale=0.06]{images/python.png}} \quad \href{https://github.com/RishavR/TTR-Generator}{\includegraphics[scale=0.08]{images/github.png}} \quad \href{https://github.com/styfeng/GenAug/tree/master/code/evaluation}{\includegraphics[scale=0.08]{images/github.png}} } \\
\multicolumn{1}{m{5cm}}{\textbf{Fluency} \break PPL, SLOR, SPELLCHECK} & \multicolumn{1}{m{6cm}}{\quad \href{https://huggingface.co/spaces/evaluate-metric/perplexity}{\includegraphics[scale=0.1]{images/hf.png}} \quad \href{https://pytorch.org/torcheval/stable/torcheval.metrics.html}{\includegraphics[scale=0.3]{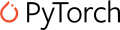}}\quad  \href{https://torchmetrics.readthedocs.io/en/stable/text/perplexity.html?highlight=perplexity}{\includegraphics[scale=0.6]{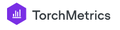}} \quad  \href{https://github.com/styfeng/GenAug/tree/master/code/evaluation}{\includegraphics[scale=0.08]{images/github.png}} } \\
\multicolumn{1}{m{5cm}}{\textbf{Semantic Content Preservation} \break BPRO} & \multicolumn{1}{m{6cm}}{\quad \href{https://huggingface.co/spaces/evaluate-metric/bertscore}{\includegraphics[scale=0.1]{images/hf.png}} \quad \href{https://pypi.org/project/bert-score/}{\includegraphics[scale=0.06]{images/python.png}} \quad \href{https://github.com/styfeng/GenAug/tree/master/code/evaluation}{\includegraphics[scale=0.08]{images/github.png}}} \\				\multicolumn{1}{m{5cm}}{\textbf{Sentiment Consistency} \break SENTSTD, SENTDIFF} & \multicolumn{1}{m{6cm}}{\quad \href{https://github.com/styfeng/GenAug/tree/master/code/evaluation}{\includegraphics[scale=0.08]{images/github.png}}} \\
\hline
\end{tabular}
\caption{Metrics for machine translation quality involved in text data augmentation.}
\label{tab:table_3}
\end{table}

\subsection{Character n-gram matches}
Word-level n-gram-based metrics, including BLEU, relied on surface form matching, and they are not appropriate for subword and character-level neural machine translation systems. If the target side language is morphologically rich, these metrics are not preferable because similar words present several forms. The character n-gram F-score (CHRF) metric is language and tokenization-independent with good correlations in human judgments. Table \ref{tab:table_4} presents the Python language implementations for this metric.

\begin{table}[h!]
\centering
\begin{tabular}{cc}
\hline
\textbf{Metric} & \textbf{Python-based Tools}\\
\hline
\multicolumn{1}{m{3cm}}{CHRF} &  \multicolumn{1}{m{5cm}}{\quad \href{https://github.com/m-popovic/chrF}{\includegraphics[scale=0.08]{images/github.png}} \quad  \href{https://huggingface.co/spaces/evaluate-metric/chrf}{\includegraphics[scale=0.1]{images/hf.png}} \quad \href{https://www.nltk.org/api/nltk.translate.chrf_score.html}{NLTK}} \\
\hline
\end{tabular}
\caption{Metric for character n-gram matches in text data augmentation.}
\label{tab:table_4}
\end{table}

\subsection{Prediction quality for classification}
Prediction quality is a set of metrics computed in the augmentation evaluation process. They are measurements that show how close the text augmentation model’s predictions are to the actual values. Table \ref{tab:table_5} shows the different metrics used to measure the prediction quality.

\begin{table}[h!]
\centering
\begin{tabular}{cc}
\hline
\textbf{Metric} & \textbf{Python-based Tools}\\
\hline
\multicolumn{1}{m{4cm}}{Accuracy} &  \multicolumn{1}{m{6cm}}{\quad \href{https://huggingface.co/spaces/evaluate-metric/accuracy}{\includegraphics[scale=0.1]{images/hf.png}} \quad \href{https://pytorch.org/torcheval/stable/generated/torcheval.metrics.MulticlassAccuracy.html}{\includegraphics[scale=0.3]{images/pytorch.png}} \quad \href{https://torchmetrics.readthedocs.io/en/stable/all-metrics.html}{\includegraphics[scale=0.6]{images/tm.png}}} \\
\multicolumn{1}{m{4cm}}{Precision} &  \multicolumn{1}{m{6cm}}{\quad \href{https://huggingface.co/spaces/evaluate-metric/precision}{\includegraphics[scale=0.1]{images/hf.png}} \quad \href{https://pytorch.org/torcheval/stable/torcheval.metrics.html}{\includegraphics[scale=0.3]{images/pytorch.png}} \quad \href{https://torchmetrics.readthedocs.io/en/stable/all-metrics.html}{\includegraphics[scale=0.6]{images/tm.png}}} \\
\multicolumn{1}{m{4cm}}{Macro-F1 Score} &  \multicolumn{1}{m{6cm}}{\quad \href{https://huggingface.co/spaces/evaluate-metric/f1}{\includegraphics[scale=0.1]{images/hf.png}} \quad \href{https://pytorch.org/torcheval/stable/torcheval.metrics.html}{\includegraphics[scale=0.3]{images/pytorch.png}} \quad \href{https://torchmetrics.readthedocs.io/en/stable/all-metrics.html}{\includegraphics[scale=0.6]{images/tm.png}}} \\
\multicolumn{1}{m{4cm}}{Recall score} &  \multicolumn{1}{m{6cm}}{\quad \href{https://huggingface.co/spaces/evaluate-metric/recall}{\includegraphics[scale=0.1]{images/hf.png}} \quad \href{https://pytorch.org/torcheval/stable/torcheval.metrics.html}{\includegraphics[scale=0.3]{images/pytorch.png}} \quad \href{https://torchmetrics.readthedocs.io/en/stable/all-metrics.html}{\includegraphics[scale=0.6]{images/tm.png}}} \\
\multicolumn{1}{m{4cm}}{AUC} &  \multicolumn{1}{m{6cm}}{\quad \href{https://huggingface.co/spaces/evaluate-metric/roc_auc}{\includegraphics[scale=0.1]{images/hf.png}} \quad \href{https://torchmetrics.readthedocs.io/en/stable/all-metrics.html}{\includegraphics[scale=0.6]{images/tm.png}}} \\
\multicolumn{1}{m{4cm}}{Jaccard similarity} &  \multicolumn{1}{m{6cm}}{\quad \href{https://torchmetrics.readthedocs.io/en/stable/all-metrics.html}{\includegraphics[scale=0.6]{images/tm.png}}} \\
\multicolumn{1}{m{4cm}}{Matthew’s correlation, \break similarity score} &  \multicolumn{1}{m{6cm}}{\quad \href{https://huggingface.co/spaces/evaluate-metric/matthews_correlation}{\includegraphics[scale=0.1]{images/hf.png}} \quad \href{https://torchmetrics.readthedocs.io/en/stable/classification/matthews_corr_coef.html?highlight=matthew}{\includegraphics[scale=0.6]{images/tm.png}}} \\
\multicolumn{1}{m{4cm}}{Confusion matrix} &  \multicolumn{1}{m{6cm}}{\quad \href{https://pytorch.org/torcheval/stable/torcheval.metrics.html}{\includegraphics[scale=0.3]{images/pytorch.png}} \quad \href{https://torchmetrics.readthedocs.io/en/stable/classification/confusion_matrix.html?highlight=confusion}{\includegraphics[scale=0.6]{images/tm.png}}} \\
\multicolumn{1}{m{4cm}}{F1 standard deviation} &  \multicolumn{1}{m{6cm}}{\quad \href{https://huggingface.co/spaces/evaluate-metric/f1}{\includegraphics[scale=0.1]{images/hf.png}}} \\
\multicolumn{1}{m{4cm}}{AUPRC} &  \multicolumn{1}{m{6cm}}{\quad \href{https://pytorch.org/torcheval/stable/torcheval.metrics.html}{\includegraphics[scale=0.3]{images/pytorch.png}} \quad \href{https://torchmetrics.readthedocs.io/en/stable/classification/average_precision.html?highlight=AUPRC}{\includegraphics[scale=0.6]{images/tm.png}}} \\
\multicolumn{1}{m{4cm}}{CTF} &  \multicolumn{1}{m{6cm}}{\quad \href{https://github.com/SaiSakethAluru/Counterfactual-fairness}{\includegraphics[scale=0.07]{images/github.png}}} \\
\multicolumn{1}{m{4cm}}{NLL} &  \multicolumn{1}{m{6cm}}{\quad \href{https://pytorch.org/docs/stable/generated/torch.nn.NLLLoss.html}{\includegraphics[scale=0.3]{images/pytorch.png}}} \\
\hline
\end{tabular}
\caption{Metrics for prediction quality for classification in text data augmentation.}
\label{tab:table_5}
\end{table}

\subsection{Datasets relationship}
The set of metrics of this category allows us to measure the following aspects. First, the relationship between the two datasets is based on Spearman’s rank. If positive correlations exist, data in datasets one and two increases. Negative correlations imply that as dataset one increases, dataset two decreases. Correlations of -1 or +1 indicate an exact monotonic relationship. Second, dataset transformation enhances the data by rescaling and transforming its features. Table \ref{tab:table_6} shows the metrics involved in this category.

\begin{table}[h!]
\centering
\begin{tabular}{cc}
\hline
\textbf{Metric} & \textbf{Python-based Tools}\\
\hline
\multicolumn{1}{m{4cm}}{Spearman’s rank correlation, Cosine similarity} &  \multicolumn{1}{m{6cm}}{\quad \href{https://huggingface.co/spaces/evaluate-metric/spearmanr}{\includegraphics[scale=0.1]{images/hf.png}} \quad \href{https://torchmetrics.readthedocs.io/en/stable/pairwise/cosine_similarity.html?highlight=cosine}{\includegraphics[scale=0.6]{images/tm.png}}} \\
\hline
\end{tabular}
\caption{Metric for a relationship between two datasets.}
\label{tab:table_6}
\end{table}

\subsection{ASR Performance}
The metric used for ASR systems is Word Error Rate (WER), which compares the predicted output and the target transcript word by word to resolve the number of differences between them. On the other hand, the Word Recognition Rate (WRR) measures the noise level of ASR transcripts. Table \ref{tab:table_7}

\begin{table}[h!]
\centering
\begin{tabular}{cc}
\hline
\textbf{Metric} & \textbf{Python-based Tools}\\
\hline
\multicolumn{1}{m{4cm}}{WER, WRR} &  \multicolumn{1}{m{6cm}}{\quad \href{https://huggingface.co/spaces/evaluate-metric/wer}{\includegraphics[scale=0.1]{images/hf.png}} \quad \href{https://torchmetrics.readthedocs.io/en/stable/text/word_error_rate.html}{\includegraphics[scale=0.6]{images/tm.png}}\quad \href{https://github.com/belambert/asr-evaluation}{\includegraphics[scale=0.07]{images/github.png}}} \\
\hline
\end{tabular}
\caption{Metric used for ASR performance systems for text augmentation techniques.}
\label{tab:table_7}
\end{table}

\subsection{Training}
The metrics of this category give insight into the behavior of the augmentation algorithms and their impact on model performance. An error rate, defined as $1-accuracy$, refers to the degree of prediction error of a model. Table \ref{tab:table_8} shows Cross-entropy loss that gives information on how correct a particular prediction is.

\begin{table}[h!]
\centering
\begin{tabular}{cc}
\hline
\textbf{Metric} & \textbf{Python-based Tools}\\
\hline
\multicolumn{1}{m{4cm}}{Cross-entropy loss} &  \multicolumn{1}{m{6cm}}{\quad \href{https://pytorch.org/docs/stable/generated/torch.nn.CrossEntropyLoss.html}{\includegraphics[scale=0.3]{images/pytorch.png}}} \\
\hline
\end{tabular}
\caption{Metric to measure prediction performance of the model after training.}
\label{tab:table_8}
\end{table}

\subsection{Pre-trained language model robustness}
The selection of a PLM and the layers for transfer learning improve the performance on a target task and prevent negative transfer. Thus, TransRate measures the transferability as the mutual information between features of target examples extracted by a PLM and their labels. From the perspective of feature representation, TransRate evaluates both the completeness and compactness of pre-trained features. A set of metrics allows the improvement of PLM. Attack accuracy is the core metric to reflect better robustness, where a larger query number indicates that the vulnerability of the target model is harder to detect. The perturbed percentage is the ratio of perturbed word numbers to the text length, and a larger percentage reveals more difficulty in successfully attacking the model. Table \ref{tab:table_9} shows metrics to select and improve the robustness of PLMs.

\begin{table}[h!]
\centering
\begin{tabular}{cc}
\hline
\textbf{Metric} & \textbf{Python-based Tools}\\
\hline
\multicolumn{1}{m{4cm}}{TransRate} &  \multicolumn{1}{m{7cm}}{\(TrR_{T_s} \rightarrow T_t(g,\in)=R(\hat{Z},\in)-R(\hat{Z},\in|Y)\)} \\
\multicolumn{1}{m{4cm}}{Attack accuracy, \break Query number, \break Perturbation Ratio} &  \multicolumn{1}{m{7cm}}{\quad \href{https://github.com/LinyangLee/BERT-Attack}{\includegraphics[scale=0.07]{images/github.png}}} \\
\hline
\end{tabular}
\caption{Metrics to select and improve robustness of pre-trained language models.}
\label{tab:table_9}
\end{table}

\subsection{Manipulated words importance}
The word importance is measured by the explainability derived from an XAI model. Subsequently, the explainability score of the word manipulated in the augmentation process reflects the final labeling process of the augmented data used to derive the soft labeling result. No computational implementation was found, in that case, the explainability score ($S_c$) is calculated as follows \(S_c = [S_c(X^1_1),\dots,S_c(X^n_m)]\).

\section{Conclusion}
This study investigates opportunities to explore and establish unified evaluation metrics categories for improving text data augmentation evaluation. The characteristics of the metrics and their applications in NLP tasks help to understand and measure text augmentation good practices. In addition, it was proposed a selection of computational tools that researchers and practitioners can implement and use. The tools introduce various strategies to measure quality augmented text data to enhance data augmentation work and make it more accessible and reproducible. Is expected that the results of this work serve as a guide for NLP researchers to spread the use and implementation of the metric categories and computational tools to start inspiring additional works in this area.

%Bibliography
\bibliographystyle{unsrt}  
\bibliography{references}

\end{document}